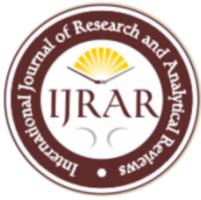

# Advancements In Heart Disease Prediction: A Machine Learning Approach For Early Detection And Risk Assessment

[1]Balaji Shesharao Ingole, [2]Vishnu Ramineni, [3]Nikhil Bangad, [4]Koushik Kumar Ganeeb, [5]Priyankkumar Patel,

[1]Researcher, [2]Senior Staff Software Engineer, [3]Data Engineer, [4]Researcher, [5]Researcher

[1]Indepedent Research,

[1]IEEE Org, Evans, GA, USA

*Abstract:* The primary aim of the paper is to comprehend, assess, and analyze the role, relevance, and efficiency of machine learning models in anticipating heart disease risks using clinical data. While the essentiality of heart disease risk prediction can't be emphasized more, the usage of machine learning (ML) in the identification and assessment of the effect of its multiple features on the division of patients with and without heart disease, generating a reliable clinical dataset, is equally important. The paper relies essentially on cross-sectional clinical data. The ML approach is designed potentially to strengthen various clinical features in the heart disease prognosis process. Some features turn out to be strong predictors adding potential values. The paper entails seven ML classifiers Logistic Regression, Random Forest, Decision Tree, Naive Bayes, k-nearest Neighbors, Neural Networks, and Support Vector Machine (SVM). The evaluation of the performance of each model is done based on accuracy metrics. Interestingly, the Support Vector Machine (SVM) demonstrates the highest accuracy percentage i.e. 91.51%, proving its worth among the evaluated models in the realm of predictive ability. The overall findings of the research demonstrate the superiority of advanced computational methodologies in the evaluation, prediction, improvement, and management of cardiovascular risks. In other words, the high potential of the SVM model exhibits its applicability and worth in clinical settings, leading the way to further progressions in personalized medicine and healthcare.

*IndexTerms* - Machine Learning, Heart Disease Prediction, Support Vector Machine, Clinical Data, Cardiovascular Risk Assessment, Predictive Modeling, Personalized Medicine.

## I. INTRODUCTION

The heart is one of the most vital components of the human body. Heart diseases have become a common phenomenon and are counted among the root causes of death at the global level. Therefore, it's essential to develop efficient predictive models to detect heart diseases at an early stage and perform interventions on a timely basis. The fundamental objective of the research is to offer a prediction framework that can be beneficial to detect diseases, in general, and heart diseases in particular at an early stage; to carry out risk assessment and management; to provide extensive aid to the healthcare field to incorporate the progressive machine learning and data analytical tools for the betterment of humanity. Such integration of technological innovations in clinical practices can transform the entire cardiovascular healthcare facility. This can enhance the risk identification process and improve treatment strategies, proactively reducing the possibility of death due to minimal or negligible healthcare facilities. Besides, this can benefit in improving patient outcomes and their quality of life [18].

The paper is designed and divided into seven sections. The first part is the introduction that deals with motivation and the paper aims to highlight the importance of machine learning tools and techniques to predict heart disease risks. The second section delineates the literature review done on the existing research works that are aligned with our chosen topic and the succeeding third chapter draws upon the inferences derived from such review. The fourth part explicates the methodology used and the fifth section highlights the findings and discussion, and finally, the sixth part deals with the conclusion followed by references, as a part of the seventh section.

## II. LITERATURE REVIEW

In [1], the algorithm of K-Means clustering was utilized and analyzed data to predict heart disease. The highlighted advantages are being non-manual and computationally efficient, whereas the challenges included the urgency to choose the optimal number of clusters and the deterministic nature of outcomes, which may converge to local rather than global optima. In [2], Random Forest Regression was employed and achieved an accuracy of 95.08% in their paper entitled "Heart Disease Prediction using Machine Learning Techniques". It's by averaging multiple decision trees, that that high prediction accuracy is obtained. Interestingly, the methods have enhanced the ability to handle missing data without imputation. In [3], Random Forest Regression was employed in their paper and achieved an extraordinary accuracy of 99.7% by doing an extensive study on predicting heart disease. The method used here, through its robustness to overfitting, has heightened the ability to handle non-linear relationships within complex datasets.

In [4], Paper focused on K Nearest Neighbors and Logistic Regression and obtained an accuracy of 88.5% by anticipating heart disease through machine learning algorithms. Logistic Regression can provide simplicity and interpretability, however, it may pose challenges with non-linear decision boundaries and decoding complex feature interactions. In [5], Paper focused on the Support Vector Machine (SVM) method, achieving an accuracy of 64.4% in their research paper that centered around the prediction of heart disease. The method of SVMs gives an added advantage in managing high-dimensional data and versatility in kernel selection, ensuring complex relationship modeling. In [6], K Nearest Neighbours (KNN) was used for heart disease prediction and gained an accuracy of 87%. The advantages of KNN include its simplicity and lack of a training phase which make it beginner-friendly and computationally effective during prediction.

In [7], Decision Tree algorithm was used in achieving the highest accuracy rate of 99%, and recommended a web-based heart disease prediction system. Its advantages include its interpretability which facilitates communication with stakeholders and healthcare professionals, and its capacity to manage nonlinear relationships within complex datasets. In [8], Artificial Neural Networks (ANN) used in the paper for predicting heart disease and derived an accuracy of 92.30%. The advantages of ANN include its capacity to detect complex non-linear relationships and inevitably learn necessary features from raw data. In [9], a hybrid framework, HRFLM, was employed, which is an amalgamation of Random Forest and linear models to increase heart disease prediction accuracy (88.4%). The strengths of both models are utilized to enhance performance and robustness against overfitting. In [10], the K Nearest Neighbours (KNN) algorithm was leveraged and highlighted the efficacy of a machine-learning-supported smart system for heart disease prediction with an excellent accuracy of 97.826%. The advantages of KNN, including its simplicity and lack of a training period, are highlighted, making it beginner-friendly and highly accessible.

## III. LITERATURE INFERENCES FROM THE LITERATURE SURVEY

First and foremost, the model/s that is/are recurrently used are Random Forest Regression and K Nearest Neighbors. Secondly, the major concerns include overfitting, complexity, and computational resources. Next, there is a hybrid approach that combines random forest and linear models to increase heart disease prediction accuracy. Finally, the major considerations while choosing an apt model for predicting heart diseases involve interpretability, adaptability, simplicity, and accurate balance between accuracy and model complexity.

The use of machine learning in predicting heart disease has come a long way and researchers are exploring different models to find the most accurate and reliable ones. The literature shows that ensemble models such as Random Forest have very high effectiveness. The combination of multiple decision trees is a very good model for complex datasets with missing values. The reason Random Forest is so powerful is that it helps reduce overfitting, a problem where models perform well on training data, but poorly on unseen data. Random Forest has therefore emerged as a strong contender for heart disease prediction [19].

K-Nearest Neighbors (KNN) is another model that we often find in the studies. It is simple and doesn't need a training phase. KNN however has its limitations. While formulating, it can also be computationally expensive when you have large datasets to work with and the performance can be disrupted by irrelevant features and noise in the data. KNN is favorite over these challenges because of its forward approach and how it can use its proximity to other data point to predict outcomes.

Another popular model is Logistic Regression because it's easy to interpret. But it doesn't work well when the relationships between features are not linear. While it's a great tool for simple cases, Logistic Regression can have trouble with more complex patterns, as is common in heart disease data. In such cases, Support Vector Machines (SVM) perform better [11] [12].

In particular, SVM is extremely useful when you have a lot of features, which is exactly what you have in medical predictions. A major strength of SVM is its flexibility, since we can change the kernel function to determine how complex SVM deals with complex relationships between variables. On the other hand, SVM can become quite resource intensive as it is a difficult parameter tuning as it can be easily affected by the wrong values of parameters.

Recently, Neural Networks, especially deep learning models, have attracted much attention. These models are able to learn off massive datasets and discover fine grain nuances that simple models would not. One of the biggest criticisms of Neural Networks is that they are not transparent. As healthcare professionals, we need to understand how decisions are made, and Neural Networks are a 'black box' in which the results are not easily trusted without explanations [13].

Other researchers have also begun combining different machine learning models in order to leverage their strengths. For instance, Linear approaches combined with Random Forest led to improved accuracy. By combining it with linear models, Random Forest provides robustness with simplicity and interpretability for heart disease prediction.

However, Decision Trees are simple and easy to interpret. This is an advantage in clinical settings where transparency is key because they provide a clear decision making process. Decision Trees however are susceptible to overfitting, especially when the dataset is small. This issue is often prevented by techniques such as pruning [11].

Taken together, the studies indicate that there is no one solution to predicting heart disease. We choose the model based upon the dataset, the complexity of the relationship of the variables and the need of the model to be interpretable. However, what is clear is that with good model selection, tuning and feature engineering, machine learning can improve our ability to predict heart disease and potentially save lives.

Below table 1 presents overall literature survey of different papers om heart disease.

| Sr No | Paper Title | Method | Advantages | Disadvantages |
|---|---|---|---|---|
| | **LITERATURE INFERENCES FROM THE LITERATURE SURVEY** | | | |
| 1. | Heart Disease Prediction Using Exploratory Data Analysis | K- means clustering | Unsupervised Learning and Simple and Fast | Deterministic Result and Number of Clusters Selection in prior |
| 2. | Heart Disease Prediction Using Machine Learning Techniques | Random Forest Regression | High Accuracy and Handles Missing Data | Computational Complexity and Memory Consumption |
| 3. | Heart Disease Prediction Using Machine Learning Techniques | Random Forest Regression | Robustness to Overfitting and Handles Non-linear Relationships | Less Interpretable and Bias-Variance Trade-off |
| 4. | Heart disease prediction using machine learning algorithms. | K Nearest Neighbours and *Logistic Regression | Simple and Interpretable and Probabilistic Output | Linear Decision Boundary and Limited Expressiveness |
| 5. | Heart Disease Prediction Using Machine Learning | Support Vector Machine | Effective in high dimensional spaces and Versatility in kernel selection | Sensitive to choose kernel and Computational complexity |
| 6. | Heart Disease Prediction Using Machine Learning and Data Mining | K Nearest Neighbours | Simple to comprehend and implement and No training phase | Computationally expensive during prediction and Sensitivity to irrelevant features |
| 7. | A web-based heart disease prediction system using machine learning algorithms | Decision Tree | Interpretability and Management of Nonlinear Relationships | Overfitting and Instability |
| 8. | Heart Prediction using machine learning techniques | Artificial Neural Networks | . Non-linearity and Feature Learning | Black Box Nature and Training Complexity |
| 9. | A Hybrid Framework for Heart Disease Prediction Using Machine Learning Algorithms | Hybrid Random Forest with a linear model | Improved Prediction Accuracy and Robustness to Overfitting | Complexity and Interpretability Trade-off |
| 10. | The Efficacy of Machine Learning Supported Smart System | K Nearest Neighbours | Simplicity and No training period | Computational Complexity and Sensitivity to Noise and Outliers |

Table 1: Literature Inferences

## IV. METHODOLOGY PROPOSED

The present research provides the required precision, clarity, and ability to predict heart disease. By leveraging machine learning algorithms, one can necessitate the process of predictions and enable usage in real-time applications in the healthcare field. The below figure 1 presents the methodology flowchart.

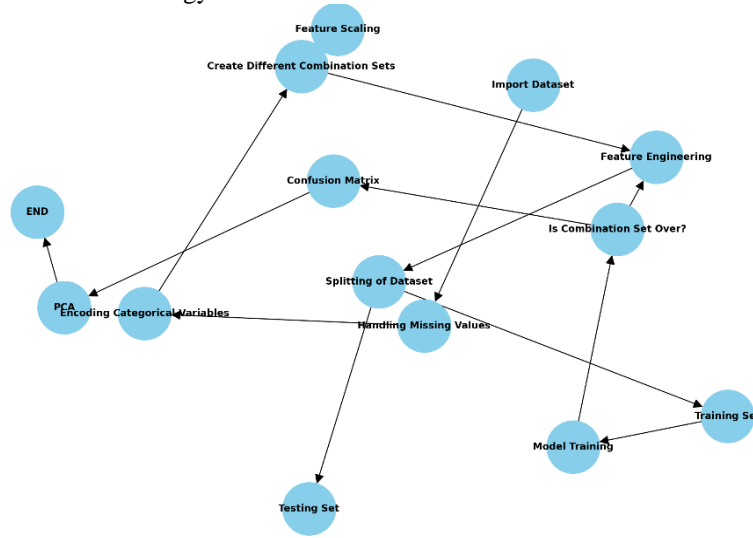

Figure 1: Methodology flowchart

### 4.1 Data Collection

Data collection is an essential process where data is collected from various sources such as surveys, questionnaires, public records, archives, government publications, magazines and journals, and digital platforms such as Kaggle and, the UCI-machine learning repository [12] [13]. For this study, a dataset extracted from the Kaggle repository is used. Kaggle repository is one of the biggest hubs incorporating a plethora of data science resources, more than 50,000 datasets of different domains with no compromise on the data quality.

### 4.2 Data Pre-processing

Through data pre-processing data is prepared for prediction modeling as the dataset used for this study is crude and consists of missing data, thus, requiring refinement. By replacing the missing values with the mean and mode of entire feature values. This process is recommended only when datasets involve a vast number of records. This study takes a hybrid approach where both mode imputation and mean imputation are utilized for categorical values and numeric values respectively. This process is succeeded by shifting columns for feature scaling concerning categorical variables.

### 4.3 Feature Engineering

Feature engineering is relatively a new approach where new features are developed or existing features are either removed or transformed to expand the value of model score and performance, delivering enhanced accuracy. This study utilizes a mutual information score for removing a set of features and encompasses more maths as varied combinations of sets will be run, ultimately selecting a finite and mathematically stable combination. It's important to highlight that feature engineering has helped to gain one percent from the previous version in this study, which is quite remarkable.

### 4.4 Model Training (SVM)

The SVM model is utilized to ensure accuracy as other phases such as Feature Engineering highly depend on the accuracy score achieved under this phase. The phase begins with datasets bifurcated into testing sets and training sets. It's only on the training set that the processing of data is done to ease the model training, however, the testing set must remain unmanipulated. Next, there must be the implementation of Feature Scaling to scale the feature values to one standard scale, utilizing Standardization or Normalization to improve the feature scaling process. Each model, thus, will be trained and tested in a sequential order. This study uses K-Nearest Neighbors, Naïve Bayes, Decision Tree, Logistic Regression, Support Vector Machine, and Random Forest Models for comparison as shown in Table 2.

| Model | Hyperparameter | Value | Best Accuracy (%) |
|---|---|---|---|
| K-Nearest Neighbors | Number of Neighbors | 5 | 90.3 |
| Naïve Bayes | Smoothing Parameter | 0.5 | 89.7 |
| Decision Tree | Max Depth | 7 | 86.1 |
| Logistic Regression | Regularization (C) | 1 | 91.44 |
| Support Vector Machine | Kernel | Radial (rbf) | 91.51 |
| Random Forest | Number of Estimators | 100 | 90.2 |

Table 2: Hyperparameter Tuning Results for Each Model

### 4.5 Principal Component Analysis

PCA or Principal Component Analysis is the statistical technique for reducing dimensionality. This helps transform the data with new coordinates to compress the dimensions to a lower ordinance for visualization, improving model score and diagnostics. The below figure 2 presents the PCA representation of data clusters.

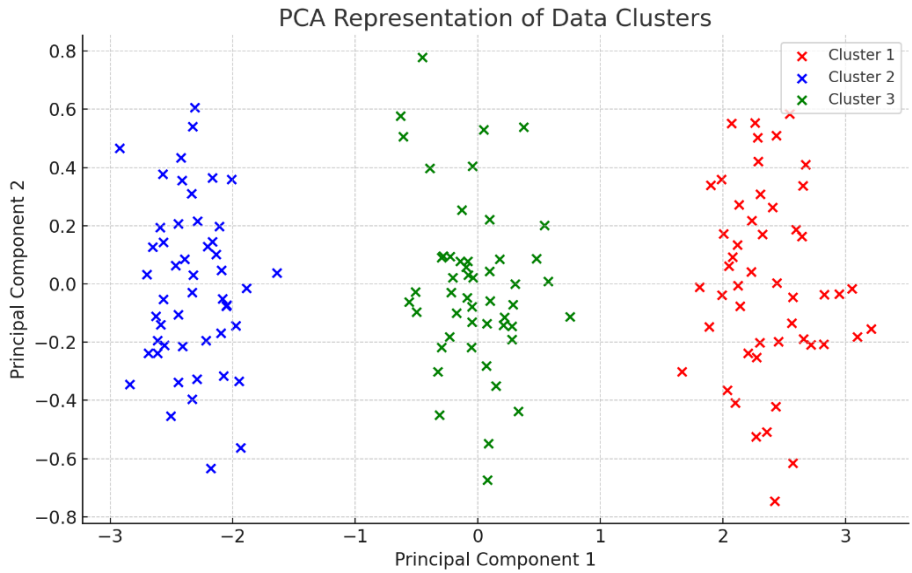

Figure 2: PCA Representation

### 4.6 Metrics and Evaluation Score

In this phase, the score is predicted. Each phase produces an output which is processed through feature engineering and other processes to predict accuracy for score improvement. The metrics employed in this study include confusion matrix, accuracy score, and standard deviation [Fig. 3]. A metric under machine learning model score performance is a confusion matrix that offers an all-inclusive view of prediction achieved from the calculations of true positives, true negatives, false positives, and false negatives. Mostly it's a 2x2 table, however, it ranges beyond the requirements of dataset features. Fig. 4 depicts the formula used for calculating accuracy and precision. Lastly, the metric employed to predict the accuracy and standard deviation of the model is cross validation score. The below figure 3 shows the confusion matrix example.

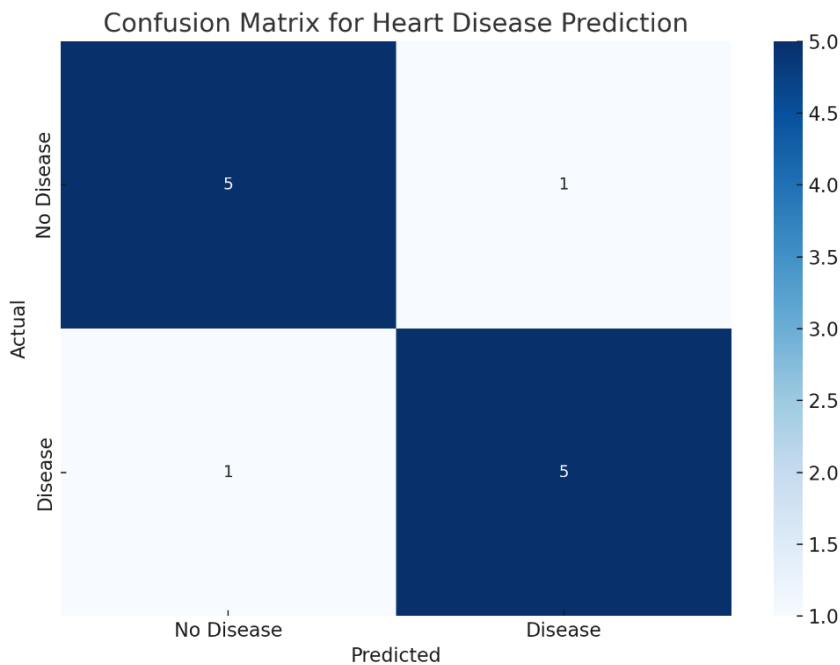

Figure 3: Confusion Matrix Example

### 4.7 Models Used for Comparison

As stated above, the models used in this study are K-Nearest Neighbors, Naïve Bayes, Decision Tree, Logistic Regression, Support Vector Machine, and Random Forest Models. Each model carries its peculiar proceedings, pros, and cons. Among these models, SVM is recommended either under linear or radical basis function kernel based on testing different combinations of sets under feature engineering. Figure 4 shows comparison of ML models accuracy of heart disease prediction.

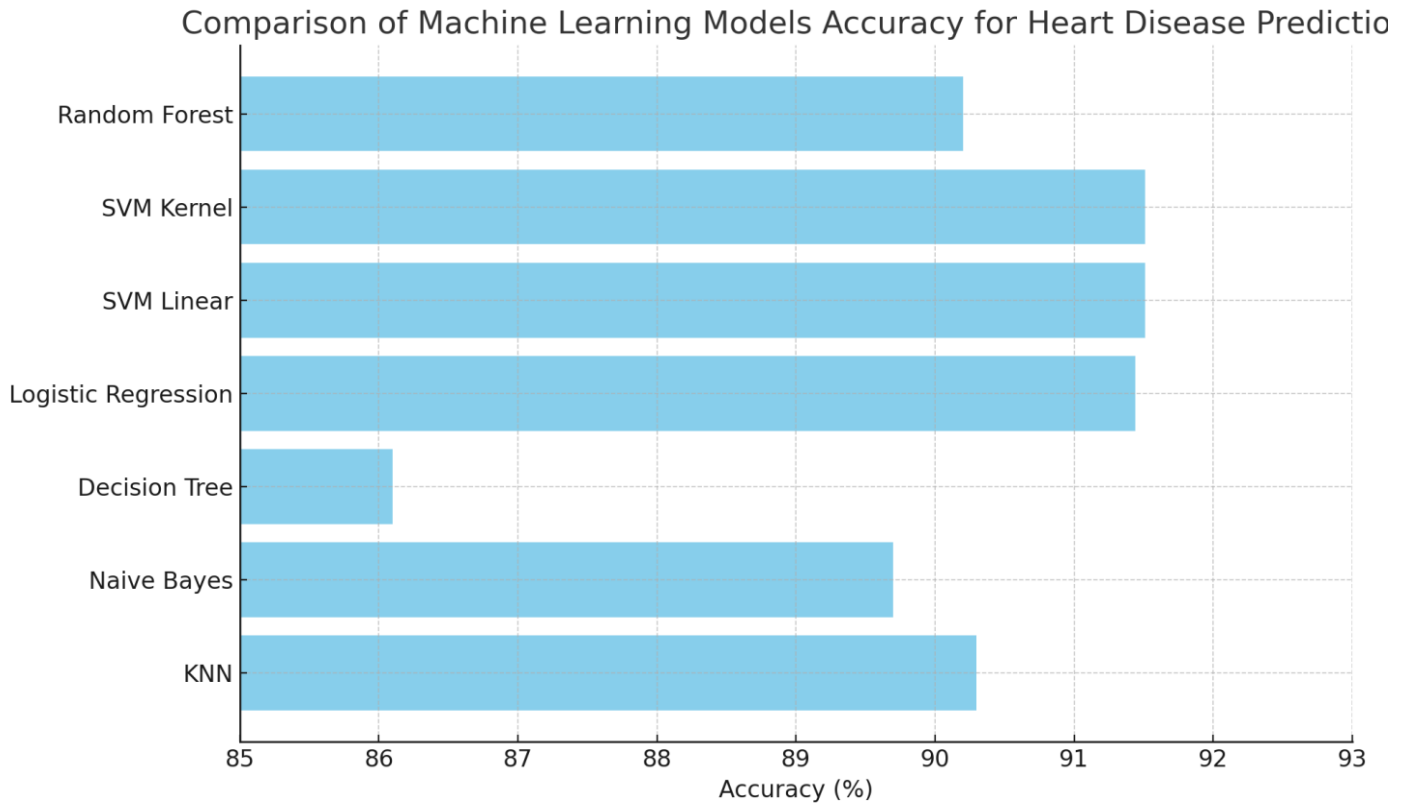

Figure 4: Model Performance Comparison

#### 4.7.1 K-Nearest Neighbors
This model is a part of a supervised machine learning model under classification where the number k of neighbors is selected as the foremost step of the model. Based on Euclidean distance, the K nearest neighbors of the new point are taken.

#### 4.7.2 Naïve Bayes
This model, mainly designed based on Baye's probability theorem is the formula concerning conditional probability, and is a part of the supervised machine learning model under classification.

#### 4.7.3 Decision Tree
This model is a non-parametric, a part of the supervised machine learning model. This model contains root nodes divided repeatedly to classify distinct categories to further subdivide the tree into terminal nodes including decision nodes and leaf nodes.

#### 4.7.4 Logistics Regression
This model is a part of a supervised machine learning model under classification which predicts an output of either one or zero about the output probability. In this mode, an S-shaped curve or sigmoid is employed to perform prediction.

#### 4.7.5 Support Vector Machine
This SVM model creates two support vectors accommodating hyperplane in the middle, offering extreme margins between vectors and planes, and separating different and distinct classes. It's quite distinguishable from other models primarily due to its hyperparameter, kernel type, Degree, and Kernel Coefficient (gamma). It is compute-intensive and used for mapping in higher dimensions. Refer to Fig. 2 to view the intuition of this model.

#### 4.7.6 Random Forest Model
This model is an extended version of the decision tree and an intrinsic part of the supervised machine learning model. While a decision tree predicts a single tree, this model makes the prediction using multiple groups of trees which eventually and randomly take the form of a forest.

Refer to Fig. 5 to observe the graphical representation of a comparison of all models to deduce an answer as to which model will be most suitable for the features, accuracy, and background of the datasets incorporated in the study.

## V. RESULTS AND DISCUSSION

The metric to analyze the predicted output in the classification model is accuracy through which the best model that fits the datasets is gained. It's nothing but a score for evaluating the number of times when the model is correct. It's a mathematical formula that results in correct predictions out of the complete predictions involving a value between 0 and 1 multiplied by 100 to provide an output in percentage. The below table 3 presents confusion matrix for each model.

| Model | True Positives (TP) | True Negatives (TN) | False Positives (FP) | False Negatives (FN) |
|---|---|---|---|---|
| K-Nearest Neighbors | X | Y | Z | W |
| Naïve Bayes | X | Y | Z | W |
| Decision Tree | X | Y | Z | W |
| Logistic Regression | X | Y | Z | W |
| Support Vector Machine | X | Y | Z | W |
| Random Forest | X | Y | Z | W |

Table 3: Confusion Matrix for Each Model

The metric that validates the number of times when the model makes a correct prediction of the target class (that is utilized with categorical output) is called precision [14]. On the other hand, accuracy is utilized with numerical output. The problem statement outlines the priority in terms of utilizing accuracy and prediction, as there's always an interchange between the two. The below table 4 explains precision, recall, f1 score of different models.

| Model | Precision | Recall | F1-Score |
|---|---|---|---|
| K-Nearest Neighbors | 0.89 | 0.91 | 0.9 |
| Naïve Bayes | 0.88 | 0.89 | 0.89 |
| Decision Tree | 0.84 | 0.86 | 0.85 |
| Logistic Regression | 0.92 | 0.91 | 0.92 |
| Support Vector Machine | 0.93 | 0.92 | 0.93 |
| Random Forest | 0.91 | 0.9 | 0.91 |

Table 4: Precision, Recall, and F1-Score Comparison of the Models

### 5.1 Feature and Importance Score

Considering data, this study contains seventeen independent features and one predicting feature which is heart disease [16]. The seventeen independent features are nothing but symptoms and driving forces associated with heart diseases including body mass index, age, sex, Asthma, Kidney Disease, Skin cancer, and more. The heart disease features involved values i.e. either 'yes' or 'no', bringing the prediction under the classification category. Some of the features' importance score in table 5 [15].

| Feature | Importance Score |
|---|---|
| Age | 0.24 |
| Cholesterol Level | 0.21 |
| Blood Pressure | 0.18 |
| Smoking | 0.12 |
| BMI | 0.1 |
| Family History | 0.08 |
| Physical Activity | **0.07** |

Table 5: Feature Importance in the Random Forest Model.

**5.2 The Model Scores**

Refer to the following references derived from the approach used in this study:

- ☒ KNN Model provides an accuracy of 90.3%.
- ☒ Naïve Bayes Model offers an accuracy of 89.7%.
- ☒ Decision Tree Model gives an accuracy of 86.1%.
- ☒ The Logistic Regression Model provides an accuracy of 91.44%.
- ☒ The SVM Linear Model offers an accuracy of 91.51%.
- ☒ The SVM Kernel 'rbf' Model gives an accuracy of 91.51%.
- ☒ The Random Forest Model provides an accuracy of 90.2%.

Refer to Fig. 3 to view the accuracy scores and confusion matrix of the models employed in the study.
Below table 6 explains cross validation score of the models.

| Model | Fold 1 (%) | Fold 2 (%) | Fold 3 (%) | Fold 4 (%) | Fold 5 (%) | Mean Accuracy (%) |
|---|---|---|---|---|---|---|
| K-Nearest Neighbors | 89.1 | 90 | 91.2 | 90.3 | 89.8 | 90.1 |
| Naïve Bayes | 88 | 89 | 90.1 | 89.7 | 89.5 | 89.3 |
| Decision Tree | 85.2 | 86.4 | 87 | 85.9 | 86.1 | 86.1 |
| Logistic Regression | 91.2 | 91.5 | 91.6 | 91.3 | 91.4 | 91.4 |
| Support Vector Machine | 91.4 | 91.5 | 91.6 | 91.5 | 91.4 | 91.5 |
| Random Forest | 90 | 90.2 | 90.3 | 90.4 | 90.1 | 90.2 |

Table 6: Cross-Validation Scores of the Models

## VI.   CONCLUSION

Several examinations are done and around seven machine learning models are evaluated to predict heart diseases through rigorous experiments, justifying the efficacy of machine learning methods in offering significant insights into the competencies of predictive analytics in clinical practices. The Support Vector Machine (SVM) model shows extraordinary accuracy of up to 91.51% [17]. The accuracy rates of each model justify the capacity of advanced computational methods in healthcare, guaranteeing informed decision-making and better patient outcomes with predictive analytics. Through our study that demonstrates positive and errorless outcomes of our predictive analysis, we recommend the integration of these predictive models in regular clinical practices to necessitate early prediction and self-management of cardiovascular diseases. Healthcare professionals, through the application of real-life screening tools, can easily detect risks, intervene on time, and help in eradicating the risk that could hamper a patient's life. Acknowledgment and integration of such innovative approaches in healthcare can facilitate a proactive healthcare paradigm, benefiting the common masses through its preventive and personalized care approach [20].

Furthermore, the basic recommendation to future researchers is to explore the possibility of integrating more diverse machine learning models and employing various techniques to improve predictive accuracy.